\documentclass{article} 
\PassOptionsToPackage{numbers,compress}{natbib}
\PassOptionsToPackage{table}{xcolor}
\usepackage[preprint]{Archive}

\usepackage{amsmath,amsfonts,bm}

\def\eqref#1{equation~\ref{#1}}

\def\1{\bm{1}}

\DeclareMathAlphabet{\mathsfit}{\encodingdefault}{\sfdefault}{m}{sl}
\SetMathAlphabet{\mathsfit}{bold}{\encodingdefault}{\sfdefault}{bx}{n}

\usepackage{url}
\usepackage{pifont}  
\usepackage{booktabs}
\usepackage{tabularx}
\usepackage[table]{xcolor}
\definecolor{archiveblue}{HTML}{245F91}
\usepackage[breaklinks=true,colorlinks=true,citecolor=archiveblue,linkcolor=archiveblue,urlcolor=archiveblue]{hyperref}

\makeatletter
\newcommand*{\myfnsymbol}[1]{%
  \ensuremath{%
    \ifcase#1\or
      \text{\ding{41}}\or
      1\or
      2\or
      3\or
      4\or
      5\or
      6\or
      7\else
      8\fi
  }%
}
\def\@fnsymbol#1{\myfnsymbol{#1}}
\def\thempfootnote{\myfnsymbol{\c@mpfootnote}}
\makeatother

\definecolor{NowWAMshade}{RGB}{240,245,250}
\usepackage{graphicx}
\title{Beyond Future Prediction: Denoising as Generative Adaptation for Robot Control}

\author{ 
\vspace{-42pt}\\
\textbf{Zanyi Wang}$^{1}$\hspace{2mm}
\textbf{Yuheng Lei}$^{2}$\hspace{2mm} 
\textbf{Dengyang Jiang}$^{3}$ \\
\textbf{Ping Luo}$^{2}$\hspace{2mm} 
\textbf{Mengdi Wang}$^{4}$\hspace{2mm}
\textbf{Zhixuan Liang}$^{4,2}$\textsuperscript{\ding{41}}\hspace{2mm}
\textbf{Shilong Liu}$^{4}$\thanks{Corresponding authors}\\ [1mm]
$^1$UC San Diego \hspace{3mm} $^2$HKU 
\hspace{3mm} $^3$HKUST \hspace{3mm} $^4$Princeton University\\[1.5mm]
\url{https://xmz111.github.io/NowWAM}
}

\begin{document}

\maketitle

\begin{figure}[h]
  \vspace{-0.7em}
    \centering
    \includegraphics[width=0.98\linewidth]{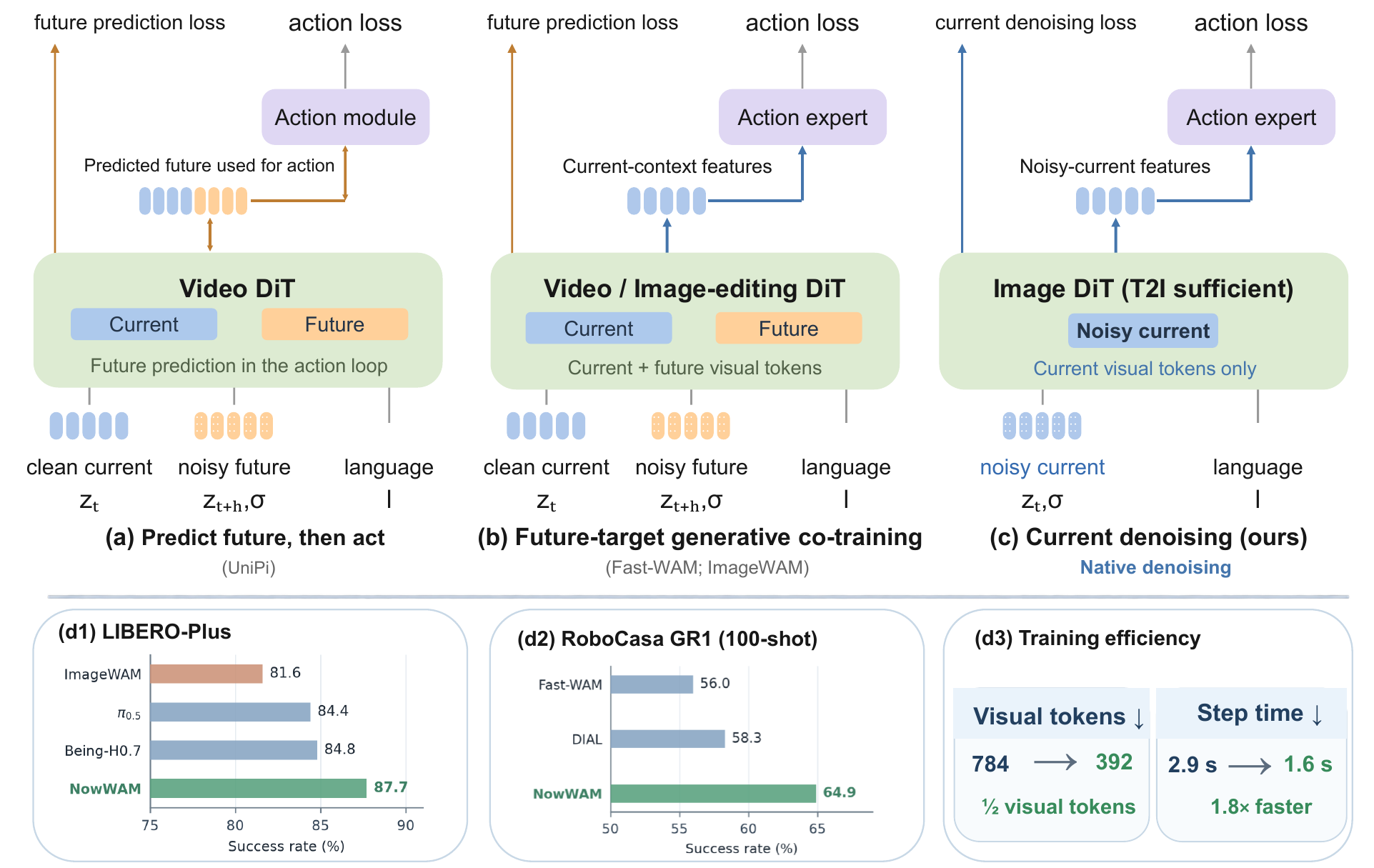}
{\small
\caption{\textbf{From future prediction to current denoising.}
(a) Future prediction is used for action prediction.
(b) Future targets are used for generative co-training.
(c) NowWAM denoises the current visual stream and predicts actions directly.
(d) NowWAM improves robustness while reducing training cost.}
\label{fig:overview}}
\vspace{5pt}
\end{figure}

\begin{abstract}
Pretrained generative Diffusion Transformers (DiTs) capture rich pixel-level visual and language-conditioned structure through large-scale image and video generation training. A growing line of robot policies builds on this generative prior through generative visual and action co-training, commonly instantiated as future visual prediction, yet recent methods increasingly move this prediction out of the inference and keep it only for co-training. This shift leaves open a more basic question, what a pretrained generative DiT actually contributes to action learning, and how this prior should be adapted for control. We introduce \textbf{NowWAM}, a future-target-free co-training formulation that denoises the current observation and predicts robot actions from the same visual stream, directly coupling the native generative objective to the action-facing representation across the denoising trajectory.
Under matched controlled settings, past and future visual targets perform comparably, whereas restricting training to the clean endpoint substantially reduces robustness. This suggests that a separate future target is not essential for generative adaptation, but the continuum of denoising states remains an effective interface for control. On LIBERO-Plus, NowWAM reaches 87.7\% with FLUX2-Klein, improving over the future-target co-training baseline by 6.1 points while halving training visual tokens (784$\rightarrow$392) and reducing step time from 2.85 s to 1.63 s, a 1.8$\times$ speedup. With the pure text-to-image Z-Image backbone, NowWAM still reaches 87.8\%, confirming that strong control adaptation does not depend on video generation or image-editing backbones. 
\end{abstract}

\section{Introduction}

Large-scale pretraining has become the foundation of modern robot policies. A representative type of work is vision-language-action models that inherit semantic representations from pretrained vision-language models and adapt them to action prediction~\citep{kim2024openvla,black2024pi_0,bjorck2025gr00t,intelligence2025pi_,qwenmanip}. Relatively speaking, another parallel line of work instead builds policies on pretrained generative Diffusion Transformers (DiTs), whose pretraining captures rich pixel-level visual and language-conditioned structure for generation. Yet how such generative priors should be transferred to control remains unclear. Existing generative policies commonly instantiate this transfer through future visual prediction, making forecasting a prevailing interface between generative pretraining and action learning~\citep{du2023learning,wu2024unleashing,ye2026world}.

However, the role of future prediction has progressively shifted in generative robot policies. As illustrated in Fig.~\ref{fig:overview}, earlier approaches place predicted futures directly inside the action loop, while more recent methods remove future generation from deployment and retain it only for generative co-training~\citep{du2023learning,wu2024unleashing,ye2026world,yuan2026fast,zhang2026imagewam}. This progression raises a basic question: if the future is no longer used for action inference, does predicting the future itself explain why generative co-training helps? Our controlled comparison offers a simple clue that under matched settings, past and future visual targets yield comparable robustness, so the benefit cannot be uniquely attributed to forward temporal semantics. This in turn motivates us to consider \emph{what a pretrained generative DiT actually contributes to action learning and how this prior should be adapted for control}.

\begin{figure*}[t]
    \centering
    \includegraphics[width=\textwidth]{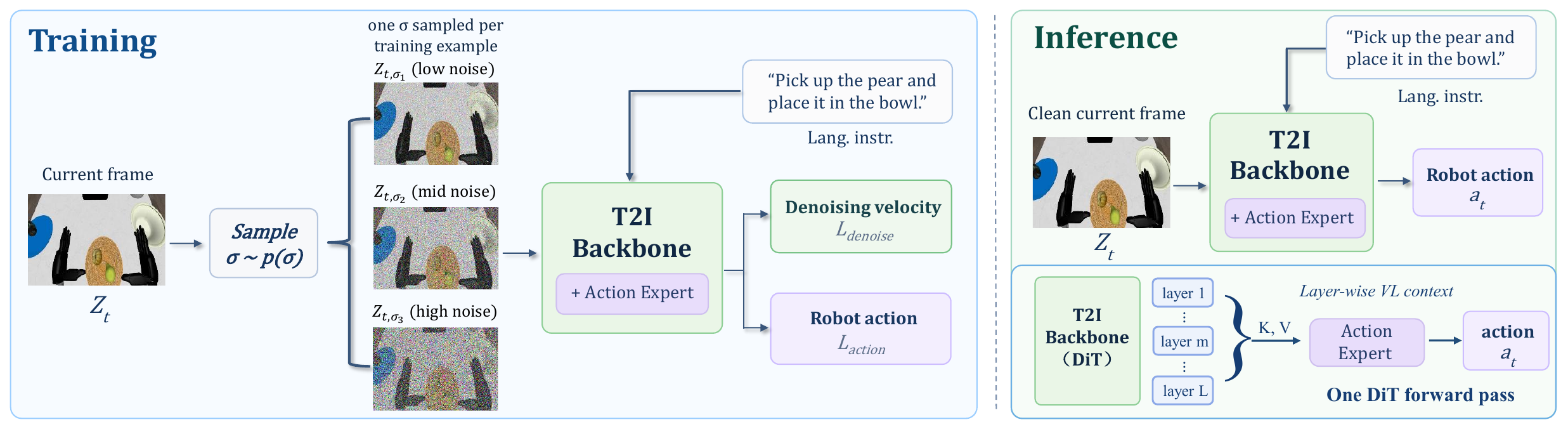}
\caption{\textbf{NowWAM.} During training, the current visual stream is sampled along the pretrained denoising trajectory and jointly supports generative prediction and action learning. At inference, the same policy operates at the clean endpoint ($\sigma=0$) in a single forward pass.}
    \label{fig:NowWAM_train_eval}
\end{figure*}

To answer this, we go back to the native denoising process of a generative DiT itself. We find that applying this denoising process directly to the visual stream used for control already provides a useful training signal for action learning, with no auxiliary target required. Motivated by this observation, we propose \textbf{NowWAM} (Fig.~\ref{fig:NowWAM_train_eval}), which removes the separate temporal target and applies denoising directly to the current visual stream used for control. This makes the same visual stream serve both generative adaptation and action learning, rather than maintaining a separate auxiliary target. The action route itself is otherwise unchanged, and the key intervention lies only in where the generative objective is applied. Importantly, this is not simply a current-frame replacement for future-target co-training. The action-facing visual stream itself is sampled along the DiT's denoising trajectory and jointly optimized for generative velocity prediction and action learning. The action expert is therefore trained with visual-language context spanning a continuum of denoising states rather than only the clean endpoint, directly coupling the native generative objective to the representation used for control. This single-stream formulation also removes the additional target tokens used by future-target co-training. At inference, the same policy simply evaluates the clean endpoint in a single forward pass, with no auxiliary visual stream or visual denoising rollout. By reducing generative adaptation to this native interface, NowWAM can further extend to a pure text-to-image DiT, showing that temporal generative pretraining is also not required.

This simpler formulation is also substantially more robust under distribution shift. On LIBERO-Plus, NowWAM achieves 87.7\% success with FLUX2-Klein and 87.8\% with Z-Image, outperforming state-of-the-art VLA and generative-policy baselines in our comparison while maintaining near-saturated performance on standard LIBERO~\citep{fei2025libero,blackforestlabs2026flux2klein,cai2025z,liu2023libero}. Its gains are concentrated on challenging visual perturbations, consistent with improved robustness of the current-state representation. The strong performance of the pure text-to-image Z-Image backbone further shows that the NowWAM formulation transfers beyond image-editing pretraining to a pure text-to-image DiT. Removing the target stream also halves the number of visual tokens during training (784$\rightarrow$392), providing a simpler and more efficient interface between generative pretraining and action learning.

Our contributions are threefold:
\begin{itemize}
    \item \textbf{We dissect how pretrained generative vision priors transfer to robot control.} Controlled analyses show that past and future targets perform comparably, while restricting adaptation to the clean endpoint of denoising process substantially reduces robustness.
    \item \textbf{We introduce NowWAM, a future-target-free generative co-training policy.} NowWAM applies current-frame denoising to the action-facing representation, coupling generative adaptation with action learning in a single stream.
    \item \textbf{We show that this simpler interface generalizes across benchmarks and architectures.} NowWAM improves performance on LIBERO-Plus and RoboCasa, extends from image-editing backbones to a pure text-to-image DiT, and halves the visual tokens used during joint training.
\end{itemize}

\section{Related Work}

\paragraph{Pretrained VLM backbones for robot control.}
Generalist robot policies increasingly adapt large-scale pretrained visual and language representations to action prediction, with modern VLA and diffusion-based foundation models scaling this paradigm across heterogeneous datasets, embodiments, tasks, and action spaces~\citep{o2024open,team2024octo,brohan2023rt,kim2024openvla,black2024pi_0,intelligence2025pi_,bjorck2025gr00t,liu2025rdt,liang2026discrete,wen2025dexvla,kim2025fine,zheng2026x,yang2026abot,qwenmanip,qwenvla,adaptdiff}. These backbones are primarily understanding-oriented, and their pretrained representations are reused for control without altering the underlying visual-language interface.

\paragraph{Generative backbones for robot control.}
Large generative models, by contrast, also learn transferable visual structure well beyond their native synthesis interface. Representations extracted from pretrained diffusion models support semantic correspondence, open-vocabulary recognition, dense matching, and geometry-aware understanding across architectures and denoising states, and have further been adapted to dense prediction tasks such as depth, geometry, and segmentation~\citep{tang2023emergent,luo2023diffusion,stracke2025cleandift,xu2023open,hedlin2023unsupervised,ke2024repurposing,fu2024geowizard,kondapaneni2024text,peebles2023scalable,he2025lotus,xu2025matters,wang2026rgb,jiang2026deforming}. These results suggest that the transferable value of pretrained representations is not tied to preserving their original output interface, motivating us to ask how generative representations should be reorganized when jointly adapted to action learning.

Pretrained generative models have been incorporated into robot control through interfaces such as generated visual subgoals, diffusion-derived visuomotor representations, joint visual-action denoising, predictive visual representations, and generative video backbones jointly adapted for control~\citep{black2024zero,deng2026robot,guo2024prediction,hu2024video,wu2024unleashing,cheang2024gr,kim2026cosmos}. Among these, future prediction, cast as a world model coupling future visual dynamics with action, is particularly prominent, spanning joint denoising of future observations and actions~\citep{ye2026world,bi2026motus,zhu2025unified,wu2024unleashing,cheang2024gr,liang2025video,kim2026cosmos,ahawam}, causal imagine-then-act inference from predicted futures~\citep{du2023learning,zhou2024robodreamer,feng2507vidar,li2026causal,bharadhwaj2024gen2act,zhao2025cot,zhang2026dreamvla}, and unified autoregressive world-action prediction~\citep{hu2024video,cen2025worldvla}, together forming the future-in-the-action-loop paradigm in Fig.~\ref{fig:overview}(a), where deployment stays coupled to future prediction. Fast-WAM~\citep{yuan2026fast} instead moves future prediction out of deployment and retains it only as a training-time generative target, with ImageWAM~\citep{zhang2026imagewam} further reducing that target to a single conditional future image, defining the training-time future-target paradigm in Fig.~\ref{fig:overview}(b). Unlike all these methods, which still rely on some form of future visual target to drive generative adaptation, NowWAM removes the future target altogether and applies the native denoising trajectory directly to the current action-facing representation, showing generative adaptation for control does not require a distinct temporal target.

\section{Method}
\label{sec:method}

\subsection{Generative Co-training with a Separate Visual Target}
\label{sec:future_cotraining}

We study robot policies that jointly adapt a pretrained generative Diffusion Transformer (DiT) and an action predictor on robot demonstrations. Let $l$ denote the language instruction, $z_t$ the latent representation of the current observation, and $a$ the action sequence. A common way to retain the pretrained generative training interface during robot adaptation is to introduce an additional visual target $\mathbf{z}_t^{+}$, typically corresponding to one or more future observations, and optimize its generative objective together with action prediction.

For a generative noise level $\sigma$, the target is sampled along the pretrained denoising trajectory,
\begin{equation}
    \mathbf{z}_{t,\sigma}^{+}
    =(1-\sigma)\mathbf{z}_t^{+}
    +\sigma\boldsymbol{\epsilon},
    \qquad
    \boldsymbol{\epsilon}\sim\mathcal{N}(0,I).
    \label{eq:future_noising}
\end{equation}

Conceptually, joint training operates on
\begin{equation}
    [\,l\mid z_t\mid \mathbf{z}_{t,\sigma}^{+}\mid a\,].
    \label{eq:future_sequence}
\end{equation}
Here and below, the bracket notation denotes the conceptual training streams rather than literal concatenation into a single transformer; the DiT--MoT interaction is described in Sec.~\ref{sec:action_readout}.
The current observation provides the visual context for action prediction, while the additional target stream provides the generative adaptation path. In the formulation we study, the action stream does not directly consume the target tokens; the target affects action learning through the shared generative backbone and its adaptation during co-training.

At deployment, the auxiliary target stream is absent. The policy therefore reduces to
\begin{equation}
    [\,l\mid z_t\mid a\,],
    \label{eq:future_eval}
\end{equation}
where action prediction depends only on the language instruction and the current observation. This separation motivates a simple question: if the additional visual stream primarily serves to adapt the pretrained generative representation during training, must this adaptation be organized around a distinct future target?

\subsection{NOWWAM}
\label{sec:NowWAM}

We introduce \textbf{NowWAM}, which applies the pretrained generative interface directly to the current visual stream instead of a separate visual target. We sample $u\sim\mathcal{U}(0,1)$ and set
\begin{equation}
    \sigma=\frac{s u}{1+(s-1)u},
    \label{eq:noise_sampling}
\end{equation}
where $s=1$ gives uniform sampling and $s<1$ shifts samples toward the low-noise end of the denoising trajectory. The current latent is then sampled as
\begin{equation}
    z_{t,\sigma}
    =(1-\sigma)z_t+\sigma\epsilon,
    \qquad
    \epsilon\sim\mathcal{N}(0,I),
    \label{eq:NowWAM_noising}
\end{equation}
and the corresponding velocity target is
\begin{equation}
    v_t^{*}=\epsilon-z_t.
    \label{eq:NowWAM_target}
\end{equation}

Training now contains only one visual stream,
\begin{equation}
    [\,l\mid z_{t,\sigma}\mid a\,].
    \label{eq:NowWAM_sequence}
\end{equation}
The same trajectory-conditioned current representation supports both generative prediction and action learning. Unlike simply replacing the future target with a separate current-frame target, NowWAM applies the pretrained generative trajectory directly to the current representation used for control.

Let $\hat{v}_t$ denote the predicted visual velocity and $\hat{a}$ the predicted robot action. We optimize
\begin{equation}
    \mathcal{L}
    =\lambda_{\mathrm{vis}}\mathcal{L}_{\mathrm{denoise}}
    +\lambda_{\mathrm{act}}\mathcal{L}_{\mathrm{action}},
    \label{eq:joint_loss}
\end{equation}
with
\begin{equation}
    \mathcal{L}_{\mathrm{denoise}}
    =\left\|
    \hat{v}_t-(\epsilon-z_t)
    \right\|_2^2.
    \label{eq:denoise_loss}
\end{equation}
Here, $\mathcal{L}_{\mathrm{action}}$ is the masked mean-squared loss on the continuous action target, following the standard action supervision used by the policy. In our main configuration, $\lambda_{\mathrm{vis}}=0.5$ and $\lambda_{\mathrm{act}}=1.0$. Invalid action dimensions and padded action tokens are excluded from the loss.

\subsection{Trajectory-Conditioned Training, Clean-Endpoint Control}
\label{sec:action_readout}

The distinction between training and deployment is central to NowWAM. During training, the action expert reads visual-language context derived from $z_{t,\sigma}$, so action learning is coupled to representations spanning a continuum of denoising states. At inference, the same representation trajectory is evaluated at its clean endpoint:
\begin{equation}
    \sigma=0,
    \qquad
    z_{t,0}=z_t,
\end{equation}
and the policy operates on the clean current observation:
\begin{equation}
    [\,l\mid z_{t,0}\mid a\,]
    =
    [\,l\mid z_t\mid a\,].
    \label{eq:NowWAM_eval}
\end{equation}

\paragraph{Backbone and action expert.}
Our policy follows a DiT--MoT structure with a pretrained generative DiT backbone and a separate action DiT. The generative backbone processes the language and visual stream and provides layer-wise visual-language context to the action expert, whose action tokens and parameters remain separate and interact with this context through masked mixed attention. In the pure text-to-image instantiation, the backbone has no conditioning image or temporal input; the current observation is the visual latent being denoised, while language provides the external conditioning. NowWAM leaves the action pathway unchanged and modifies only the visual adaptation interface: during training, the current latent is sampled along the denoising trajectory, while deployment uses its clean endpoint. The action expert is therefore trained on visual-language representations across generative states, while inference uses only the clean current observation in a single forward pass.

\section{Experiments}

\subsection{Experimental Setup}
\label{sec:exp_setup}

\paragraph{Benchmarks.} We evaluate NowWAM in three complementary settings. Standard LIBERO~\citep{liu2023libero} serves as an in-distribution reference, where strong pretrained policies are already close to saturation, testing whether removing the separate future target preserves nominal manipulation capability. RoboCasa GR1 Tabletop~\citep{nasiriany2024robocasa,bjorck2025gr00t} evaluates few-shot transfer to a different simulator and task family; we use the 24 public ID tasks with 100 target-task demonstrations per task and 50 evaluation episodes per task, for 1,200 episodes in total. LIBERO-Plus~\citep{fei2025libero} is our primary robustness benchmark, covering seven categories of visual and environmental perturbations.

\paragraph{Evaluation protocol.}
Main benchmark results follow the full evaluation protocol for each benchmark, including all 10,030 LIBERO-Plus episodes, and use the final EMA checkpoints for NowWAM. For controlled analyses, we evaluate non-EMA checkpoints on the same fixed 1,923-episode LIBERO-Plus subset to make the larger ablation suite tractable. Within each controlled comparison, all training and evaluation settings are held fixed except for the stated intervention. We therefore use the main results to compare final policy performance and the controlled analyses to isolate individual design choices.

\paragraph{Backbones and training.}
We use FLUX.2-Klein-4B~\citep{blackforestlabs2026flux2klein} as our primary generative backbone and additionally evaluate NowWAM with Z-Image-6B~\citep{cai2025z}, a pure text-to-image DiT, to further test whether the formulation depends on image-editing or video-generation pretraining. Unless otherwise stated, the generative backbone and action expert are jointly optimized.

\paragraph{Baselines.} We compare against representative generalist VLA and diffusion-based policies, including $\pi_0$ and $\pi_{0.5}$~\citep{black2024pi_0,intelligence2025pi_}, OpenVLA-OFT~\citep{kim2025fine}, X-VLA~\citep{zheng2026x}, ABot-M0~\citep{yang2026abot}, and GR00T~\citep{bjorck2025gr00t}, as well as recent generative and world-action policies including LingBot-VA~\citep{li2026causal}, Motus~\citep{bi2026motus}, Cosmos-Policy~\citep{kim2026cosmos}, Fast-WAM~\citep{yuan2026fast}, and ImageWAM~\citep{zhang2026imagewam}. We additionally include StarVLA/StarVLA-OFT~\citep{community2026starvla}, Being-H0.7~\citep{luo2026being}, RLDX-1~\citep{kim2026rldx}, and DIAL~\citep{chen2026dial} where reported for the corresponding benchmark.

\subsection{In-Distribution and Few-Shot Manipulation}
\label{sec:indist_fewshot}

Before studying robustness under distribution shift, we first evaluate whether NowWAM preserves nominal manipulation performance and transfers beyond LIBERO with limited target-task data. Table~\ref{tab:id_robocasa} reports standard LIBERO and RoboCasa GR1 Tabletop results.

\begin{table*}[h]
\centering
\caption{\textbf{Standard LIBERO and RoboCasa GR1 manipulation.}}
\label{tab:id_robocasa}
\small
\setlength{\tabcolsep}{4pt}
\renewcommand{\arraystretch}{1.10}

\begin{minipage}[t]{0.57\textwidth}
\vspace{0pt}
\centering
\textbf{(a) Standard LIBERO}\par\smallskip
\begin{tabularx}{\linewidth}{
    @{}l
    *{5}{>{\raggedleft\arraybackslash}X}
    @{}
}
\toprule
Method & Spatial & Object & Goal & Long & Avg. \\
\midrule
$\pi_{0.5}$ & 98.8 & 98.2 & 98.0 & 92.4 & 96.9 \\
X-VLA & 98.2 & 98.6 & 97.8 & 97.6 & 98.1 \\
LingBot-VA & 98.5 & 99.6 & 97.2 & \textbf{98.5} & \textbf{98.5} \\
Motus & 96.8 & 99.8 & 96.6 & 97.6 & 97.7 \\
Fast-WAM & 98.2 & \textbf{100.0} & 97.0 & 95.2 & 97.6 \\
ImageWAM & 97.2 & 99.2 & \textbf{98.8} & 98.4 & 98.4 \\
\midrule
\rowcolor{NowWAMshade}
\textbf{NowWAM} & \textbf{99.5} & \textbf{100.0} & 97.5 & 96.5 & 98.4 \\
\bottomrule
\end{tabularx}
\end{minipage}
\hfill
\begin{minipage}[t]{0.39\textwidth}
\vspace{0pt}
\centering
\textbf{(b) RoboCasa GR1 Tabletop}\par\smallskip
\begin{tabularx}{\linewidth}{
    @{}l
    *{2}{>{\raggedleft\arraybackslash}X}
    @{}
}
\toprule
Method & Demos & SR (\%) \\
\midrule
GR00T-N1.6 & 1000 & 47.6 \\
StarVLA-OFT & 1000 & 48.8 \\
ABot-M0 & 1000 & 58.3 \\
RLDX-1 & Full & 58.7 \\
DIAL & 100 & 58.3 \\
Fast-WAM & 100 & 56.0 \\
\midrule
\rowcolor{NowWAMshade}
\textbf{NowWAM} & 100 & \textbf{64.9} \\
\bottomrule
\end{tabularx}
\end{minipage}

\vspace{2pt}
\begin{flushleft}
\footnotesize
External methods use their respective training recipes; demonstration counts refer to target RoboCasa GR1 data. Our RoboCasa evaluation follows the public 24-task ID protocol with 50 episodes per task.
\end{flushleft}
\end{table*}

\paragraph{In-distribution manipulation.}
Standard LIBERO is close to saturation for strong pretrained policies. NowWAM reaches 98.4\% average success, matching the future-target generative baseline and remaining competitive with the strongest reported policies. Thus, removing the separate future target does not compromise nominal manipulation performance.

\paragraph{Few-shot cross-benchmark transfer.}
RoboCasa provides a more challenging transfer setting under a different simulator and task family. Using only 100 target-task demonstrations per task, NowWAM reaches 64.9\% success over 1,200 evaluation episodes. Under the same 100-shot target-data budget, this improves over DIAL (58.3\%) and Fast-WAM (56.0\%), providing evidence that the NowWAM formulation transfers beyond LIBERO. Methods trained with larger target-task budgets are included in Table~\ref{tab:id_robocasa} for reference.

\begin{table}[h]
\centering
\caption{\textbf{Success rates (\%) on LIBERO-Plus across seven perturbation categories.}}
\label{tab:libero_plus}
\small
\setlength{\tabcolsep}{3pt}
\renewcommand{\arraystretch}{1.18}
\begin{tabularx}{\linewidth}{
    @{}l
    *{8}{>{\raggedleft\arraybackslash}X}
    @{}
}
\toprule
& \multicolumn{7}{c}{Perturbation category}
& \textbf{Overall} \\
\cmidrule(lr){2-8}
Method & Camera & Robot & Lang. & Light & Bkg. & Noise & Layout & SR (\%) \\
\midrule
$\pi_0$
& 13.8 & 6.0 & 58.8 & 85.0 & 81.4 & 79.0 & 68.9 & 53.6 \\
$\pi_{0.5}$
& 78.4 & \textbf{73.6} & 80.8 & 96.2 & 94.1 & 89.0 & 84.5 & 84.4 \\
StarVLA
& 52.5 & 49.8 & 88.5 & 95.7 & \textbf{95.7} & 73.0 & 76.9 & 74.1 \\
OpenVLA-OFT
& 56.4 & 31.9 & 79.5 & 88.7 & 93.3 & 75.8 & 74.2 & 69.6 \\
ABot-M0
& 60.4 & 67.9 & 86.4 & 96.2 & 91.6 & 86.4 & 82.6 & 80.5 \\
Cosmos-Policy
& 75.8 & 63.3 & 81.7 & 96.5 & 88.9 & 92.7 & 82.2 & 82.2 \\
Being-H0.7
& 82.0 & 59.0 & 82.8 & 97.8 & 90.0 & 93.5 & \textbf{88.5} & 84.8 \\
\midrule
Fast-WAM
& 15.5 & 43.4 & 67.1 & 79.3 & 52.4 & 39.3 & 60.0 & 49.5 \\
ImageWAM (FLUX2-Klein-4B)
& 77.7 & 48.3 & \textbf{89.9} & 97.5 & 86.0 & 95.2 & 81.8 & 81.6 \\
\midrule
\rowcolor{NowWAMshade}
\textbf{NowWAM} (FLUX2-Klein-4B)
& \textbf{88.6} & 70.5 & 87.5 & \textbf{98.3} & 93.0 & 97.4 & 82.4 & 87.7 \\

\rowcolor{NowWAMshade}
\textbf{NowWAM} (Z-Image-6B)
& 84.9 & 72.5 & 87.6 & 96.7 & 94.0 & \textbf{97.5} & 85.3 & \textbf{87.8} \\
\bottomrule
\end{tabularx}
\end{table}

\subsection{Robustness under Distribution Shift}
\label{sec:robustness}

We next evaluate on LIBERO-Plus, our primary robustness benchmark, which stresses policies with seven categories of visual and environmental perturbations over 10,030 episodes and is substantially more discriminative than the near-saturated standard LIBERO setting. NowWAM reaches 87.7\% success with FLUX2-Klein, improving the future-target generative baseline from 81.6\% by 6.1 points. The gains are concentrated on challenging visual shifts, including camera, robot, and background perturbations, while language performance remains comparable, indicating that NowWAM primarily improves the robustness of the current visual representation. With the pure T2I Z-Image backbone, NowWAM further reaches 87.8\%, demonstrating that the formulation also transfers to a pure text-to-image backbone. 

\begin{figure*}[h]
    \centering
    \includegraphics[width=\textwidth]{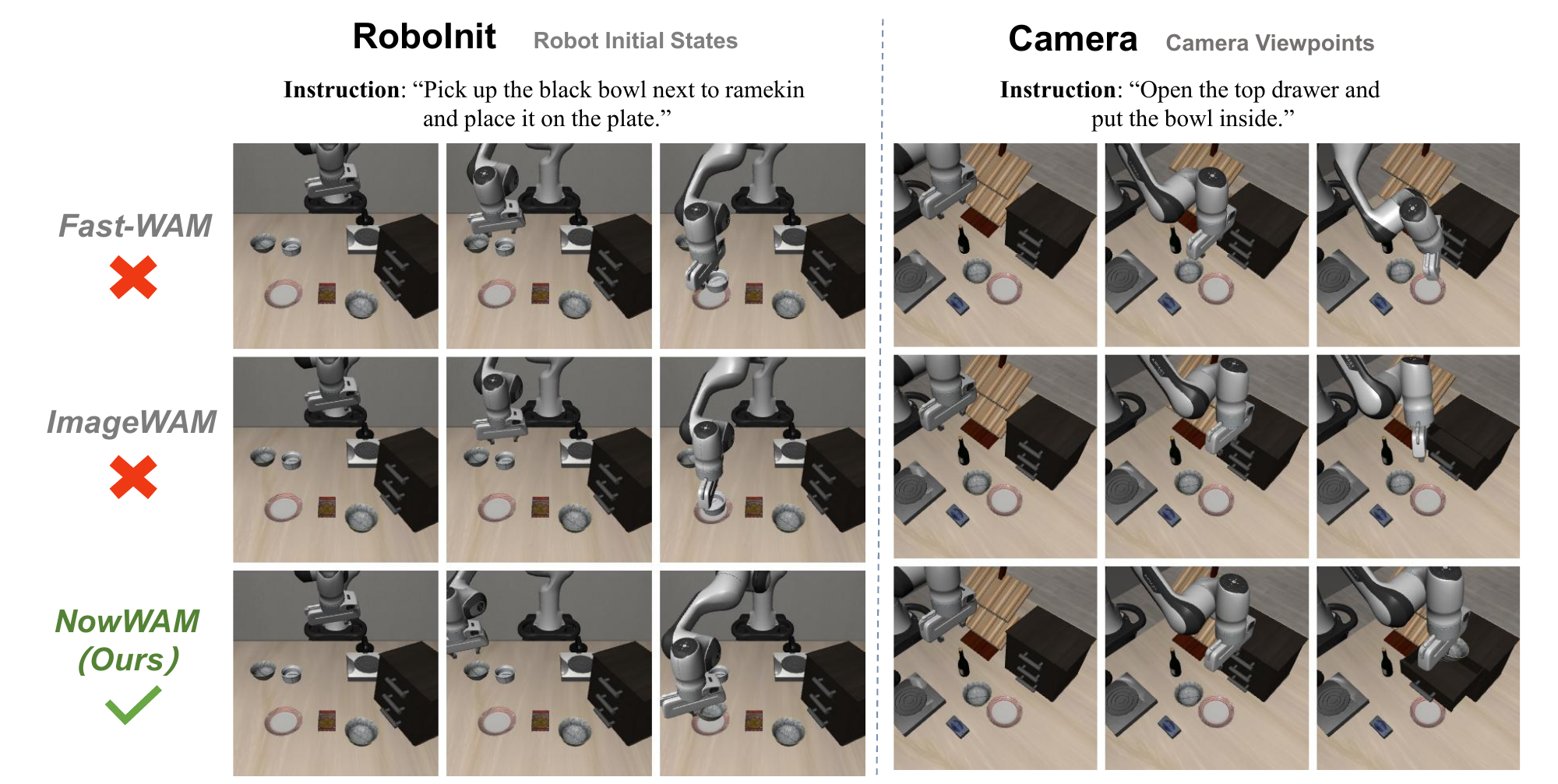}
    \caption{\textbf{Qualitative rollouts under distribution shift.}
    We show representative trajectories under RobotInit and camera-view perturbations in LIBERO-Plus.
    Fast-WAM and ImageWAM exhibit distinct grounding and execution failures, whereas NowWAM completes the corresponding tasks.
    These examples qualitatively illustrate the robustness gains observed in Table~\ref{tab:libero_plus}.}
    \label{fig:qualitative_rollouts}
\end{figure*}

Figure~\ref{fig:qualitative_rollouts} visualizes representative execution failures under two challenging perturbations. Under RoboInit, the future-target baselines misidentify the grasp target and fail to transfer the instructed black bowl, whereas NowWAM successfully grounds, grasps, and places the target bowl. Under the camera-view perturbation, the baselines fail to push open the top drawer and therefore cannot complete the required interaction, while NowWAM successfully opens the drawer and completes the full placement sequence.

\subsection{Ablations and Analysis}
\label{sec:ablations}

\begin{table}[h]
\centering
\caption{\textbf{All rows are strictly matched; only the stated intervention changes.}
Results use non-EMA checkpoints on the same fixed 1,923-episode subset.}
\label{tab:ablations}
\begin{minipage}[t]{0.35\textwidth}
\vspace{0pt}
\centering
\small
\textbf{(a) Generative initialization}\par\smallskip
\begin{tabular*}{\linewidth}{@{\extracolsep{\fill}}llr@{}}
\toprule
Init. & Visual loss & SR \\
\midrule
Random & Edit & 58.14 \\
Pretrained & Edit & 83.05 \\
Random & None & 50.55 \\
Pretrained & None & 75.81 \\
\bottomrule
\end{tabular*}
\end{minipage}\hfill
\begin{minipage}[t]{0.28\textwidth}
\vspace{0pt}
\centering
\small
\textbf{(b) Auxiliary visual target}\par\smallskip
\begin{tabular*}{\linewidth}{@{\extracolsep{\fill}}lr@{}}
\toprule
Target & SR \\
\midrule
Current target ($t$) & 79.10 \\
Future ($t+16$) & 82.89 \\
Future ($t+32$) & 83.70 \\
Past ($t-16$) & 83.80 \\
\bottomrule
\end{tabular*}
\end{minipage}\hfill
\begin{minipage}[t]{0.30\textwidth}
\vspace{0pt}
\centering
\small
\textbf{(c) Denoising adaptation}\par\smallskip
\begin{tabular*}{\linewidth}{@{\extracolsep{\fill}}lr@{}}
\toprule
Training condition & SR \\
\midrule
Uniform & 83.46 \\
Low-noise shifted & 84.97 \\
No visual loss & 78.73 \\
Clean only & 77.48 \\
\bottomrule
\end{tabular*}
\end{minipage}
\end{table}

We conduct controlled experiments on LIBERO-Plus to distinguish three design choices in generative adaptation for control: generative initialization, the temporal semantics of the visual target, and training along the denoising trajectory. Table~\ref{tab:ablations} summarizes the results. All rows follow the identical experimental contract described in Sec. \ref{sec:exp_setup}, with only the stated intervention changed. All analyses use non-EMA checkpoints.

\paragraph{Generative initialization.}
Pretrained initialization substantially improves robustness both with and without a generative objective (Table~\ref{tab:ablations}a). With the Edit objective, pretraining raises success from 58.14\% to 83.05\%; even without the visual objective, it improves performance from 50.55\% to 75.81\%. This large gain without visual supervision shows that the pretrained generative backbone already provides a strong robustness prior, while the additional generative objective further improves adaptation on top of this initialization. These results establish generative pretraining as a major source of robustness while leaving open how that prior should be adapted to control.

\paragraph{Auxiliary visual targets.}
Changing only the temporal target shows that past and future targets perform comparably (Table~\ref{tab:ablations}b). Future targets at $t+16$ and $t+32$ reach 82.89\% and 83.70\%, while the past target at $t-16$ reaches 83.80\%. Thus, forward temporal prediction is not uniquely responsible for the benefit of the auxiliary generative target. The current-target control retains the same two-stream formulation with a separate auxiliary target, and is therefore distinct from NowWAM's single-stream current denoising.

\paragraph{Denoising trajectory.}
Although NowWAM consumes clean observations at inference, training only at the clean endpoint reaches 77.48\%, substantially below training along the denoising trajectory (Table~\ref{tab:ablations}c). Here, \emph{Clean only} fixes $\sigma=0$ while retaining the visual objective, whereas \emph{No visual loss} retains noisy-current training but removes visual supervision. Uniform sampling reaches 83.46\%, while shifting the distribution toward lower noise further improves success to 84.97\%. Removing the visual denoising objective reduces success to 78.73\%. Thus, neither clean-endpoint training nor noisy-current exposure alone explains the gain. Effective adaptation therefore requires coupling action learning to the pretrained generative objective across the denoising trajectory, rather than only at the clean endpoint.

Taken together, these analyses separate three factors in generative adaptation for control. Pretrained generative initialization provides a strong robustness prior, while past and future auxiliary targets perform similarly, indicating that forward temporal semantics are not uniquely privileged. At the same time, both clean-only training and noisy-current training without visual supervision perform substantially worse than trajectory-conditioned denoising. These results suggest that the key ingredient is not future prediction itself, but retaining generative supervision on the action-facing representation across the native denoising trajectory.

\begin{table*}[h]
\centering
\caption{\textbf{Training efficiency.}
Same $2\times$H200 setup with global batch size 64. Both methods use FLUX2-Klein-4B as the backbone.}
\label{tab:training_efficiency}
\vspace{7pt}
\small
\setlength{\tabcolsep}{8pt}
\renewcommand{\arraystretch}{1.12}

\begin{tabular*}{\textwidth}{
    @{\extracolsep{\fill}}
    l
    l
    l
    l
    l
}
\toprule
Method
& Training proxy
& Vis.\ tokens $\downarrow$
& Step time (s) $\downarrow$
& Peak mem. (GiB)\ $\downarrow$ \\
\midrule

ImageWAM 
& Future image
& 784
& 2.85
& 78.9 \\

\textbf{NowWAM}
& \textbf{Current denoising}
& \textbf{392 \;(50\% fewer)}
& \textbf{1.63 \;(1.8$\times$ faster)}
& \textbf{74.3 \;(5.8\% lower)} \\

\bottomrule
\end{tabular*}

\begin{flushleft}

\end{flushleft}
\end{table*}

\subsection{Training Efficiency}
\label{sec:efficiency}

We finally compare the training cost of the two formulations under the same $2\times$H200 setup with global batch size 64. Both measurements use the same FLUX2-Klein-4B backbone and action expert. Text embeddings and VAE latents are precomputed and cached, so the reported step time and peak memory measure joint DiT--action training only and exclude text/VAE encoding. The difference therefore comes from the visual training interface: future-target co-training processes both the current reference and a separate target stream (784 visual tokens), whereas NowWAM keeps only the current stream (392 tokens).

As shown in Table~\ref{tab:training_efficiency}, removing the second visual stream cuts step time from 2.85s to 1.63s ($1.8\times$) and peak memory from 78.9 to 74.3GiB. These gains come directly from the NowWAM formulation, without additional acceleration.

\subsection{Attention Visualization}

\begin{figure}[h]
        \centering
        \includegraphics[width=\linewidth]{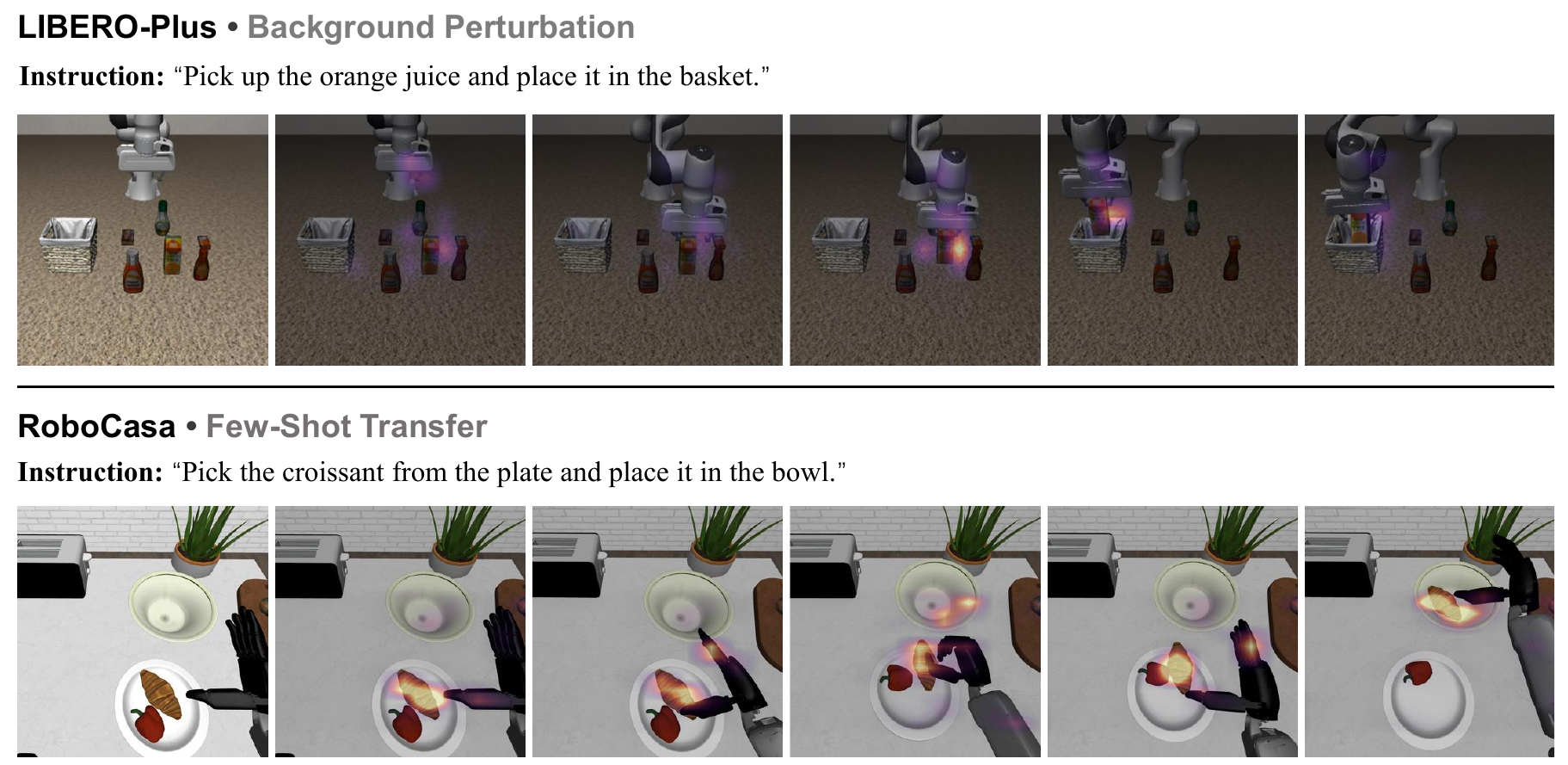}
    \caption{\textbf{Task-conditioned attention across robustness and transfer settings.}
    Each row starts with the RGB observation, followed by attention maps over the rollout.}
    \label{fig:attention}
\end{figure}
Figure~\ref{fig:attention} visualizes task-conditioned attention over successful rollouts on LIBERO-Plus and RoboCasa. Across execution stages, attention focuses on task-relevant objects and interaction regions, qualitatively illustrating the representations learned by NowWAM.

\section{Conclusion}

Pretrained generative DiTs provide a rich source of visual and language-conditioned structure, yet how their native generative training interface should be adapted to robot control remains unclear. Existing approaches commonly organize this adaptation around future visual prediction, motivating us to ask whether forward temporal semantics are essential, or whether the generative trajectory itself can serve as a more direct interface to action learning. Our controlled studies show that past and future visual targets perform comparably, while restricting adaptation to the clean endpoint substantially reduces robustness, suggesting that future prediction is not uniquely privileged whereas the denoising trajectory remains important. Building on this finding, we introduced \textbf{NowWAM}, which applies native denoising directly to the current visual stream and removes the separate future-target branch. Across LIBERO-Plus and RoboCasa, NowWAM improves robustness and few-shot transfer, extends from image-editing to pure text-to-image backbones, and reduces training cost while retaining competitive in-distribution manipulation performance. Future modeling may still be useful for explicit dynamics or long-horizon planning. In the manipulation settings studied here, however, generative pretraining can be effectively adapted through a simpler, current-centric interface without requiring a distinct future target.

\section*{Acknowledgements}
We thank Google's TPU Research Cloud (TRC) program for granting us access to Cloud TPUs.

\bibliography{Archive}
\bibliographystyle{Archive}


\end{document}